\documentclass[conference]{IEEEtran}
\IEEEoverridecommandlockouts
\usepackage{cite}
\usepackage{amsmath,amssymb,amsfonts}
\usepackage{graphicx}
\usepackage{textcomp}
\usepackage{xcolor}
\usepackage[T1]{fontenc}
\usepackage{xspace}
\usepackage{booktabs}
\usepackage[misc]{ifsym}
\usepackage{algorithm}
\usepackage{algpseudocode}
\usepackage{multirow}
\usepackage{adjustbox}
\def\BibTeX{{\rm B\kern-.05em{\sc i\kern-.025em b}\kern-.08em
    T\kern-.1667em\lower.7ex\hbox{E}\kern-.125emX}}
\begin{document}

\title{Self-Explainable Multi-Label Graph Neural Network for Correlated Evidence Attribution\\
}


\author{
\IEEEauthorblockN{
Yingqi Feng$^1$, Yufei Tang$^1$, Min Shi$^2$, and Xingquan Zhu$^1$
}
\IEEEauthorblockA{
$^1$Department of Electrical Engineering and Computer Science, Florida Atlantic University, Boca Raton, FL, USA \\
$^2$School of Computing and Informatics, University of Louisiana at Lafayette, Lafayette, LA, USA \\
\{yfeng2016, tangy, xzhu3\}@fau.edu, min.shi@louisiana.edu
}
}


\maketitle

\newcommand{\modelname}{\texttt{SEMGNN}\xspace}

\begin{abstract}
Multi-label graph learning 
intends to capture the intrinsic complexity of real-world applications, where one sample is often related to multiple groups or consists of multiple objects. To date, a handful of multi-label graph learning methods exist, but none of them integrate training-time interpretation capability. 
While post-hoc graph explainers have been developed, they do not explicitly model label-dependent evidence sharing in multi-label graph learners, especially when label pairs are weakly or negatively associated. As a result, post-hoc approaches may miss how evidence should be shared or separated across different labels. This paper advances a new end-to-end self-explainable multi-label graph neural network (\modelname), which aims to simultaneously classify multi-labeled nodes and identify edges significantly contributing to each target node \textit{w.r.t.} predicted labels. 
Different from post-hoc methods, \modelname jointly learns a predictor and a sparse edge-mask explainer within a unified framework and training objective. 
Label-label correlations are used to improve multi-label node classification and 
enhance individual 
label explanations, so that different labels of a node can be supported by distinct yet coherent structural and/or correlated evidence. Experiments and comparisons on synthetic and real-world multi-label networks, in social networking, entertainment, and life sciences, show that \modelname achieves competitive or improved predictive performance while providing more faithful and compact label-conditioned explanations.
Our code is available at: https://github.com/yfeng77/SEMGNN. 
\end{abstract}

\begin{IEEEkeywords}
Multi-label graph learning, Graph neural networks, Explainable AI, Graph explainability, Representation learning
\end{IEEEkeywords}

\section{Introduction}

Graph neural networks (GNNs) have become a dominant paradigm for learning from graph-structured data~\cite{wu2020comprehensive}, achieving strong performance by jointly modeling node attributes and relational dependencies. In many real-world graphs, a node may belong to multiple semantic categories, such as users with multiple interests, proteins with multiple functions, or entities with multiple roles, leading to multi-label node classification (MLNC)~\cite{wang2020multi}. Compared with single-label settings, MLNC is more challenging because labels are often imbalanced and correlated, and different labels of the same node may be supported by overlapping or distinct structural evidence~\cite{zhao2023multi,zhou2024multi}. Therefore, beyond predicting multiple labels, a multi-label graph learner should also explain how graph evidence is attributed to individual and correlated labels.

\begin{figure*}[t]
\centering
\includegraphics[width=0.9\textwidth]{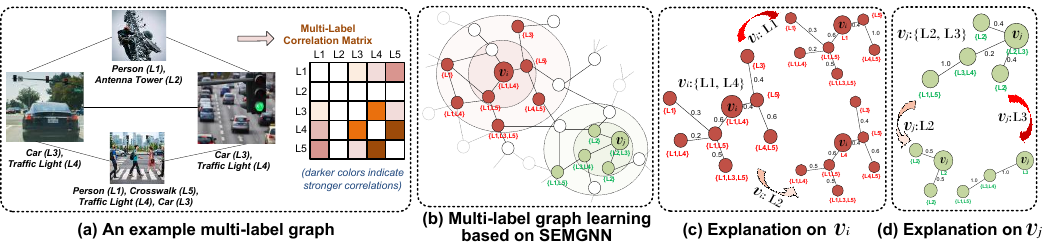}
\caption{Motivation of correlated evidence attribution in multi-label graph learning. From left to right: (a) a multi-label graph where each node is associated with multiple labels and some labels are more strongly correlated than others; (b) \modelname identifies important edges for multi-label prediction; and (c)--(d) label-conditioned explanations show that different labels of the same node may rely on distinct yet partially shared/correlated structural evidence.}
\label{fig1}
\end{figure*}

Despite the strong predictive performance of GNNs, deploying multi-label graph predictors in high-stakes domains raises a key question: \textit{Why does the model predict a particular label for a target node?} Existing GNN explanation methods mostly focus on single-label prediction~\cite{kakkad2023survey}, and many are post-hoc methods that explain a fixed predictor after training~\cite{yuan2022explainability}. This separation may weaken the alignment between prediction and explanation~\cite{ragno2025faithful}. More importantly, direct adaptations to MLNC tend to fall into two extremes: explaining each label independently may miss evidence shared by positively related labels, whereas producing a single explanation for all predicted labels can obscure label-specific rationales, especially when labels are weakly or negatively associated. This creates a distinct attribution problem in MLNC: explanations should be label-conditioned, and their overlap should adapt to label-relatedness rather than being either ignored or forced. As shown in Fig.~\ref{fig1}, a useful multi-label explanation should identify label-specific evidence while allowing statistically related labels to share support when appropriate.

To address these challenges, we propose \modelname, an end-to-end self-explainable multi-label GNN for correlated evidence attribution. Unlike post-hoc explainers, \modelname jointly learns prediction and explanation within a unified framework. It uses a shared multi-label prediction backbone and an intrinsic edge-scoring module to generate sparse explanatory subgraphs. To model label dependency, \modelname incorporates label correlations in both prediction and explanation: a label-correlation residual branch supports multi-label prediction, while a label-aware edge scorer produces label-conditioned masks guided by correlation-aware label representations. This design enables different labels of the same node to be explained by distinct but potentially overlapping evidence, rather than forcing all labels to share one explanation.

We summarize our main contributions as follows:
\begin{itemize}
\item We formulate correlated evidence attribution as a label-conditioned explanation problem for MLNC and propose \modelname, an end-to-end self-explainable framework for joint prediction and attribution.

\item We design a label-correlation-aware attribution mechanism that incorporates label dependencies into both prediction and explanation, enabling positively related labels to share explanatory evidence while preserving distinct rationales for weakly or negatively associated labels.

\item We develop a unified sufficiency--necessity--sparsity training objective and validate \modelname on synthetic and real-world multi-label graphs, showing competitive or improved prediction performance together with faithful, compact, and label-conditioned explanations.
\end{itemize}

\section{Related Work}
\subsection{Multi-label Node Classification}
Early multi-label learning methods often train one binary classifier per label~\cite{boutell2004learning,zhang2013review}, but largely ignore label dependencies. Recent GNN-based methods use shared graph encoders, such as GCN, GraphSAGE, and GAT~\cite{kipf2016semi,hamilton2017inductive,velivckovic2017graph}, with multi-label prediction heads and explicit label-relation modeling. Representative methods include ML-GCN, which models node--label and label--label correlations~\cite{shi2020multi}; LIP, which propagates higher-order label influence~\cite{sun2025multi}; and CorGCN, which preserves both label distinctiveness and correlation under ambiguous features and topology~\cite{bei2025correlation}. These methods primarily focus on predictive performance and provide limited support for interpretability.

\subsection{Post-hoc GNN Explanations}
Post-hoc GNN explainers interpret a trained predictor after model training~\cite{yuan2022explainability,ma2026post}. GNNExplainer learns instance-specific soft masks over graph structures and node features~\cite{ying2019gnnexplainer}, while PGExplainer trains a parameterized explainer to predict edge selection probabilities across instances~\cite{luo2020parameterized}. Other representative methods include GraphSVX~\cite{duval2021graphsvx}, PGM-Explainer~\cite{vu2020pgm}, SubgraphX~\cite{yuan2021explainability}, FlowX~\cite{gui2023flowx}, and recent logic-based methods~\cite{ragno2025faithful}. However, most of them are designed for single-label explanations and are learned separately from the predictor, making them less suitable for label-conditioned and correlation-aware evidence attribution in MLNC.

\subsection{Self-Explainable GNNs}
Self-explainable GNNs integrate explanation mechanisms into model training so that prediction and explanation are learned jointly~\cite{kakkad2023survey}. SE-GNN explains a target node through interpretable nearest labeled nodes~\cite{dai2021towards}, ProtGNN uses prototype-based reasoning for graph-level explanation~\cite{zhang2022protgnn}, and GSAT learns stochastic attention masks to identify task-relevant graph structures during training~\cite{miao2022interpretable}. Built-in interpretability is promising for graph learning; however, self-explainability does not guarantee faithfulness~\cite{christiansen2023faithful}, and existing self-explainable GNNs are not designed for a multi-label setting with label-specific structural explanations.

\section{Method}
\subsection{Problem Formulation}
\label{sec:problem}
Let $\mathcal{G}=(\mathcal{V},\mathcal{E})$ be an attributed graph with $N=|\mathcal{V}|$ nodes, $E=|\mathcal{E}|$ edges, node features $\mathbf{X}\in\mathbb{R}^{N\times d}$, and adjacency matrix $\mathbf{A}\in\{0,1\}^{N\times N}$. $\mathcal{V}_{\mathrm{lab}}\subseteq\mathcal{V}$ denotes the labeled nodes. For each labeled node $v_i\in\mathcal{V}_{\mathrm{lab}}$, its label is a multi-hot vector $\mathbf{y}_i\in\{0,1\}^{C}$ over the label set $\mathcal{C}=\{1,\dots,C\}$, and all labels are collected in $\mathbf{Y}\in\{0,1\}^{N\times C}$. Given $(\mathcal{G},\mathbf{X},\mathbf{Y})$, our goal is to learn a predictor $f_{\theta}$ that, for each target node $v_i$, outputs logits $\mathbf{z}_i\in\mathbb{R}^{C}$ and label probabilities $\hat{\mathbf{y}}_i=\sigma(\mathbf{z}_i)$, where $\sigma(\cdot)$ is the element-wise sigmoid. The predicted label set is:
\begin{equation}
\hat{\mathcal{C}}_i=\{c\in\mathcal{C}\mid \hat{y}_{ic}\ge \tau_{\mathrm{pred}}\},
\end{equation}
and if no label exceeds the threshold $\tau_{\mathrm{pred}}$, we keep the highest-scoring label.

Beyond prediction, we study \textit{correlated evidence attribution} for MLNC. For a target node $v_i$ and a predicted label $c\in\hat{\mathcal{C}}_i$, the goal is to extract a compact explanatory edge set $\mathcal{E}^{\star}_{i,c}\subseteq\mathcal{E}^{(K)}_i$ from the $K$-hop computation subgraph of $v_i$. Unlike single-label explanation, different labels of the same node may rely on distinct evidence, while statistically correlated labels may share part of their support. Therefore, a useful explanation should satisfy three requirements: 
(1) \textit{Faithfulness}: preserving the evidence that supports the prediction and exposing performance degradation when important edges are removed; 
(2) \textit{Compactness}: remaining sparse and easy to interpret; and 
(3) \textit{Correlation-aware label relevance}: explanations should remain label-specific, but allow related labels to share evidence when supported by label correlations. These requirements motivate an end-to-end self-explainable formulation that couples MLNC with label-conditioned and correlation-aware edge attribution.

\begin{figure*}[t]
\centering
\includegraphics[width=0.80\textwidth]{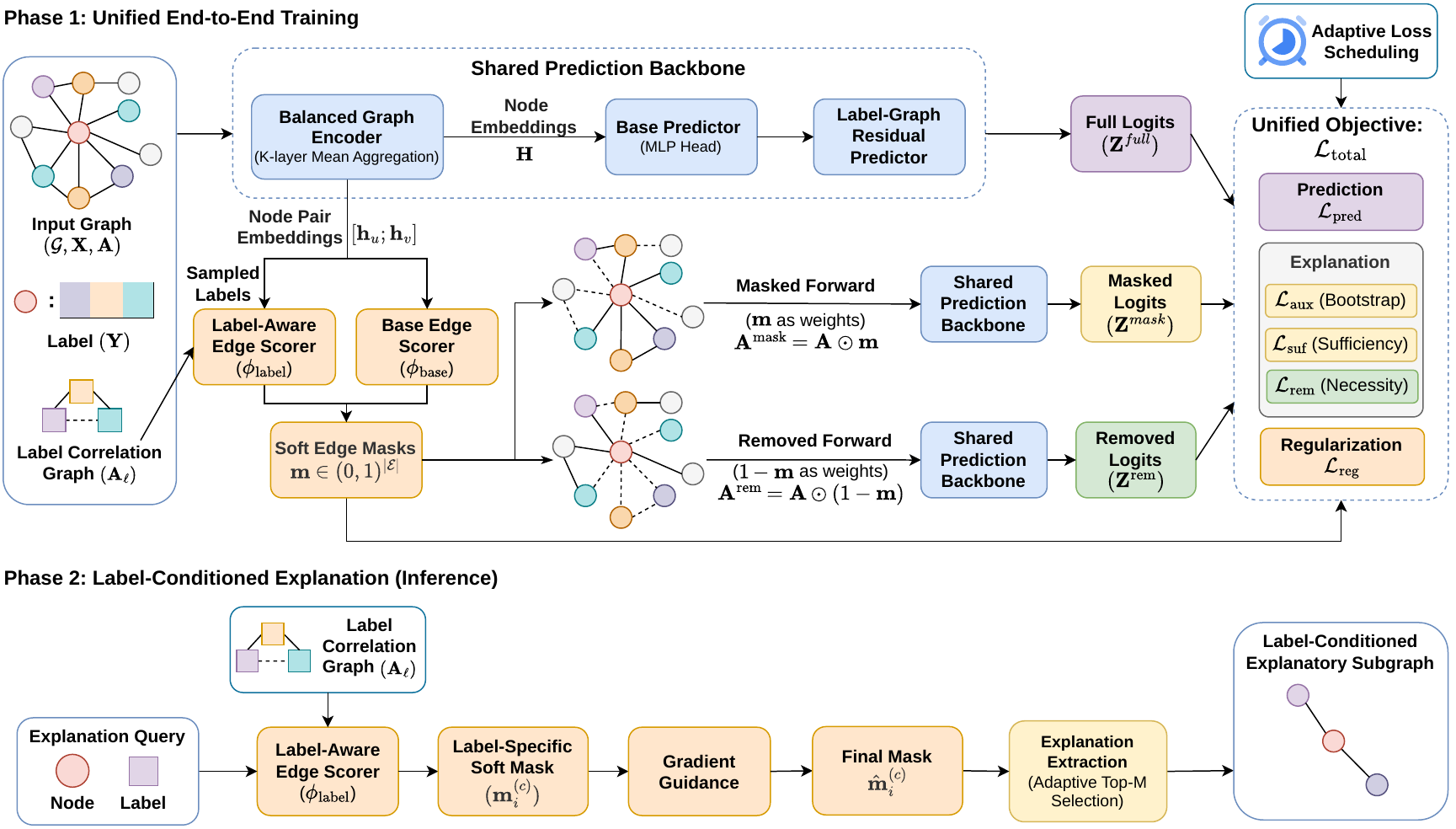}
\caption{Overview of \modelname. Phase 1 jointly learns multi-label node classification and intrinsic edge masking with a shared prediction backbone and unified end-to-end objectives. Phase 2 answers a label-conditioned attribution query by producing a label-specific mask, fusing it with gradient-based necessity scores, and extracting a compact explanatory subgraph, accounting for correlated evidence attribution.}
\label{fig}
\end{figure*}

\subsection{\modelname: Architecture Overview}
\label{sec:semgnn_overview}

\modelname is an end-to-end self-explainable framework for correlated evidence attribution in MLNC, shown in Fig.~\ref{fig}. Unlike post-hoc pipelines, it learns prediction and attribution jointly, which promotes alignment between the predicted labels and the generated label-conditioned explanations. The framework contains three main components:

\begin{itemize}
\item \textbf{Shared Prediction Backbone:}
\modelname shares the same prediction backbone across full, masked, and removed forward passes. It combines a balanced message-passing encoder with a multi-label prediction head augmented by a label-graph residual branch.

\item \textbf{Intrinsic Edge Masking:}
\modelname learns a soft edge mask $\mathbf{m}\in(0,1)^{|\mathcal{E}|}$ for masked and removed forward passes. The mask is regularized to be sparse, near-binary, and non-degenerate, making the model self-explainable by construction.

\item \textbf{Correlation-aware Evidence Attribution Module:}
\modelname uses a label-aware edge scorer to produce label-conditioned edge importance scores. By propagating label representations on the label-correlation graph, the scorer allows related labels to share attribution signals while preserving label-specific explanations.
\end{itemize}

All modules are optimized under a unified objective. During inference, for a query $(v_i,c)$ with $c\in\hat{\mathcal{C}}_i$, \modelname scores edges in the $K$-hop computation subgraph of $v_i$, refines the learned label-conditioned soft mask with gradient-based necessity scores, and extracts a compact explanatory subgraph via adaptive Top-$M$ selection.

\subsection{Shared Prediction Backbone}
\label{sec:predictor}

The full, masked, and removed forward passes in \modelname share the same prediction backbone. This shared design ensures that attribution learning is tied to the same decision function used for prediction, rather than to a separate surrogate network.

\subsubsection{Balanced graph encoder}
We encode node features with a balanced GraphSAGE-style mean encoder. At each layer, a node representation is updated by combining its current representation with the mean representation of its neighbors. This mean aggregation reduces the dominance of high-degree neighbors and yields more balanced message passing across nodes with heterogeneous degrees. Let $\mathbf{H}^{(K_{\mathrm{enc}})}$ denote the output of the $K_{\mathrm{enc}}$-layer encoder. To preserve input information, we further add a residual projection from the raw features:
\begin{equation}
\mathbf{H}
=\mathbf{H}^{(K_{\mathrm{enc}})}
+
\alpha_{\mathrm{skip}}
\mathrm{Proj}_{\mathrm{skip}}(\mathbf{X}),
\end{equation}
where $\mathbf{H}$ is the final node representation used by the predictor and edge scorer.

\subsubsection{Correlation-aware prediction head}
The encoder output $\mathbf{H}$ is first mapped by a two-layer MLP to obtain base multi-label logits. Since labels in MLNC are often correlated, we further introduce a label-correlation residual branch to provide label-dependent decision signals. Let $\mathbf{Y}_{\mathrm{tr}}\in\{0,1\}^{N_{\mathrm{tr}}\times C}$ denote the multi-hot label matrix of the training nodes. To avoid validation/test leakage, the label-correlation graph is constructed only from training labels:
\begin{equation} 
\mathbf{S}=\mathbf{Y}_{\mathrm{tr}}^{\top}\mathbf{Y}_{\mathrm{tr}}, \qquad (\mathbf{A}_{\ell})_{ab} = \frac{S_{ab}}{S_{aa}+S_{bb}-S_{ab}+\epsilon}. 
\end{equation}
For $a\neq b$, $(\mathbf{A}_{\ell})_{ab}$ measures the Jaccard-style correlation between labels $a$ and $b$, while diagonal/self-loop connections are handled during label-GCN propagation. We propagate learnable label embeddings $\mathbf{E}_{\ell}\in\mathbb{R}^{C\times d_{\ell}}$ over $\mathbf{A}_{\ell}$ using a lightweight GCN to obtain correlation-aware label representations $\widetilde{\mathbf{E}}_{\ell}$.
The final logits combine the base predictor with the gated label-correlation residual:
\begin{equation} 
\mathbf{Z} = \mathrm{MLP}_{\mathrm{pred}}(\mathbf{H}) + \beta_{\mathrm{rel}} \left( \mathrm{Proj}(\mathbf{H})\widetilde{\mathbf{E}}_{\ell}^{\top} +\mathbf{b} \right),
\end{equation}
where $\beta_{\mathrm{rel}}\in[0,1]$ is a learnable gate. This residual design preserves the stability of the base predictor while allowing correlated labels to share predictive information. It also provides correlation-aware decision signals for later label-conditioned evidence attribution.

\subsection{Edge Scoring Strategy}
\label{sec:edge_scoring}


During training, \modelname learns a soft edge mask for masked and removed forward passes, and during inference, \modelname answers a query $(v_i,c)$ with a label-specific soft mask refined by gradient-based necessity scores.

\subsubsection{Training: Soft Mask Learning}
\label{sec:edge_scoring_train}

Let $\mathbf{H}$ denote the node representations from the shared prediction backbone. For each edge $(u,v)\in\mathcal{E}$, a base edge scorer produces a label-agnostic score, while a label-aware edge scorer produces, for label $c$, a label-conditioned score to capture label-specific structural evidence:
\begin{equation}
r^{\mathrm{base}}_{uv}=\phi_{\mathrm{base}}([\mathbf{h}_u;\mathbf{h}_v]),\qquad
\Delta r^{(c)}_{uv}=\phi_{\mathrm{label}}([\mathbf{h}_u;\mathbf{h}_v],c).
\end{equation}

Directly modeling all labels at every training step is unnecessary and expensive in multi-label classification. We therefore sample a small label subset $\mathcal{S}\subseteq\mathcal{C}$ and aggregate their label-conditioned scores into a single training mask. In the reported setting, labels are sampled uniformly. The aggregated score and resulting soft mask are:
\begin{equation}
r_{uv}
=
r^{\mathrm{base}}_{uv}
+
\alpha_{\mathrm{label}}\cdot
\frac{1}{|\mathcal{S}|}
\sum_{c\in\mathcal{S}}
\Delta r^{(c)}_{uv},
\qquad
m_{uv}=\sigma\!\left(\frac{r_{uv}}{\tau_{\mathrm{mask}}}\right),
\end{equation}
where $\alpha_{\mathrm{label}}\in[0,1]$ is a learnable gate and $\tau_{\mathrm{mask}}$ is a temperature parameter. The learned mask is used in the masked forward pass, while $(1-\mathbf{m})$ is used in the removed pass. Together with the full pass, they encourage sufficiency and necessity.

\subsubsection{Inference: Label-Conditioned Scoring with Gradient Guidance}
\label{sec:edge_scoring_infer}

During inference, \modelname explains a specific query $(v_i,c)$ in the $K$-hop computation subgraph around $v_i$. It first computes a label-specific soft mask:
\begin{equation}
\mathbf{m}_{i}^{(c)}
=
\sigma\!\left(
\frac{\mathbf{r}_{i}^{\mathrm{base}}+\alpha_{\mathrm{label}}\,\Delta \mathbf{r}_{i}^{(c)}}{\tau_{\mathrm{mask}}}
\right),
\end{equation}
which allows different labels of the same node to focus on different edges. To account for label dependency, the label-aware scorer propagates the label representations over a label correlation graph before producing $\Delta \mathbf{r}_{i}^{(c)}$, so correlated labels can share part of the explanatory evidence. To further capture necessity, we fuse the queried-label mask with gradient-based scores:
\begin{equation}
\hat{\mathbf{m}}_{i}^{(c)}
=
(1-\lambda_g)\mathbf{m}_{i}^{(c)}
+
\lambda_g \mathrm{GradScore}(v_i,c).
\end{equation}
where $\lambda_g\in[0,1]$ controls the contribution of gradient guidance. The final explanation is obtained by adaptive Top-$M$ selection over $\hat{\mathbf{m}}_{i}^{(c)}$.

\subsection{Explanation Extraction}
\label{sec:extraction}

For a target node $v_i$, \modelname generates explanations only for predicted positive labels $c\in\hat{\mathcal{C}}_i$. Explanations are extracted from the $K$-hop computation subgraph $\mathcal{G}_{i}^{(K)}$ by selecting an adaptive number of undirected edge groups.

Since the computation graph is represented with directed edges, we first merge the two directions of the same connection into one undirected edge group, yielding the set $\mathcal{U}_{i}^{(K)}$ with size $G_i=|\mathcal{U}_{i}^{(K)}|$. Let $\mathbf{s}_{i}^{(c)}$ denote the final queried-label scores on these undirected edge groups, induced from the fused edge scores $\hat{\mathbf{m}}_{i}^{(c)}$. We then set the adaptive explanation budget and extract the explanation as
\begin{equation}
\begin{aligned}
M_i
&= \min\!\Big(
\max(M_{\min},\lfloor \rho G_i \rfloor),
M_{\max}
\Big),\\
\mathcal{E}^{\star}_{i,c}
&= \mathrm{Top}_{M_i}\!\big(\mathbf{s}^{(c)}_i\big).
\end{aligned}
\end{equation}

This adaptive extraction keeps explanations compact while matching the complexity of the local computation subgraph.

\subsection{End-to-End Training Objectives}
\label{sec:objective}

\modelname is trained end-to-end to jointly optimize prediction and intrinsic explanation. 
The objective contains a prediction loss, three explanation losses, and a mask regularization term:
\begin{equation}
\begin{aligned}
\mathcal{L}_{\mathrm{total}}
=&\
\underbrace{\mathcal{L}_{\mathrm{pred}}}_{\text{prediction}}
+
\underbrace{
\lambda_{\mathrm{suf}}(t)\mathcal{L}_{\mathrm{suf}}
+\lambda_{\mathrm{rem}}(t)\mathcal{L}_{\mathrm{rem}}
+\lambda_{\mathrm{aux}}(t)\mathcal{L}_{\mathrm{aux}}
}_{\text{explanation}}
\\
&+
\underbrace{
\lambda_{\mathrm{reg}}(t)\mathcal{L}_{\mathrm{reg}}
}_{\text{regularization}} .
\end{aligned}
\end{equation}
The first term is used for prediction, the next three terms train the explanation module, and the last term regularizes the learned edge mask.

\subsubsection{Prediction Loss}
\label{sec:pred_loss}

We optimize the full-graph predictor using focal binary cross-entropy(BCE) with label-wise positive weights estimated from training-label frequencies. The loss is averaged over all training nodes and labels, and is used as the prediction term $\mathcal{L}_{\mathrm{pred}}$ in the overall objective.

\subsubsection{Explanation Objectives}
\label{sec:explain_losses_new}

The explanation module is trained to satisfy both sufficiency and necessity. 
Let $\mathbf{P}^{\mathrm{full}}=\sigma(\mathbf{Z}^{\mathrm{full}})$, 
$\mathbf{P}^{\mathrm{mask}}=\sigma(\mathbf{Z}^{\mathrm{mask}})$, and 
$\mathbf{P}^{\mathrm{rem}}=\sigma(\mathbf{Z}^{\mathrm{rem}})$ denote the prediction matrices from the full, masked, and removed forward passes, respectively. 
We define the confidence weight as 
$w^{\mathrm{conf}}_{ic}=2|p^{\mathrm{full}}_{ic}-0.5|$, so that explanation supervision emphasizes labels with more confident full-graph predictions.

\paragraph{Sufficiency: Full-to-masked consistency}
We encourage the masked forward pass to preserve the full-graph prediction:
\begin{equation}
\mathcal{L}_{\mathrm{suf}}
=\frac{1}{|\mathcal{V}_{\mathrm{train}}|C}
\sum_{v_i\in\mathcal{V}_{\mathrm{train}}}
\sum_{c=1}^{C}
w^{\mathrm{conf}}_{ic} 
\cdot
\ell_{\mathrm{bce}}
\Big(
z^{\mathrm{mask}}_{ic},
\operatorname{sg}(p^{\mathrm{full}}_{ic})
\Big),
\end{equation}
where $\operatorname{sg}(\cdot)$ denotes stop-gradient and $\ell_{\mathrm{bce}}$ denotes BCE on logits.

\paragraph{Necessity: Removal supervision}
To capture necessity, we penalize confident predictions that remain after important edges are removed:
\begin{equation}
\mathcal{L}_{\text{rem}}
=
\frac{
\sum_{v_i\in\mathcal{V}_{\mathrm{train}}}\sum_{c=1}^{C}
\mathbb{I}\!\big[p^{\mathrm{full}}_{ic}>\tau_{\text{conf}}\big]\,
p^{\mathrm{rem}}_{ic}
}{
\sum_{v_i\in\mathcal{V}_{\mathrm{train}}}\sum_{c=1}^{C}
\mathbb{I}\!\big[p^{\mathrm{full}}_{ic}>\tau_{\text{conf}}\big] + \epsilon
}.
\end{equation}

\paragraph{Bootstrap: Auxiliary masked-target supervision}
To stabilize early training, we additionally supervise the masked forward pass with observed labels:
\begin{equation}
\begin{aligned}
\mathcal{L}_{\mathrm{aux}}
=\frac{1}{|\mathcal{V}_{\mathrm{train}}|C}
\sum_{v_i\in\mathcal{V}_{\mathrm{train}}}
\sum_{c=1}^{C}
w^{\mathrm{conf}}_{ic} \cdot
\ell_{\mathrm{wbce}}
\Big(
z^{\mathrm{mask}}_{ic},
y_{ic};
w_c^{\mathrm{pos}}
\Big),
\end{aligned}
\end{equation}
where $\ell_{\mathrm{wbce}}$ denotes weighted BCE on logits.

\subsubsection{Regularization}
\label{sec:reg_losses_new}

We regularize the soft edge mask $\mathbf{m}\in(0,1)^{|\mathcal{E}|}$ to be compact yet non-degenerate. To avoid over-penalizing high-degree neighborhoods, we match $\mathbf{m}$ to a degree-adaptive target $\mathbf{t}\in(0,1)^{|\mathcal{E}|}$, where each $t_e$ is obtained by log-degree scaling of the endpoints of edge $e$ and clipped into $[t_{\min},t_{\max}]$:
\begin{equation}
\begin{aligned}
\mathcal{L}_{\text{reg}}
=&\
\|\mathbf{m}-\mathbf{t}\|_2^2
+
\lambda_{\mathrm{bin}}\cdot \frac{1}{|\mathcal{E}|}\sum_{e\in\mathcal{E}} m_e(1-m_e)
\\
&+
\eta\cdot \mathrm{ReLU}\!\big(a_{\min}-\max_{e\in\mathcal{E}} m_e\big)^2,
\end{aligned}
\end{equation}
where the second term encourages near-binary masks and the last term prevents trivial collapse.

\subsubsection{Adaptive Three-Stage Coefficient Scheduling}
\label{sec:schedule}

To stabilize end-to-end training, we use an automatic three-stage coefficient schedule. At each epoch, we update exponential moving averages of model confidence and sufficiency loss, and use them to monotonically adjust the loss coefficients. 

In Stage~0 (bootstrap), we keep $\lambda_{\text{aux}}(t)>0$ while setting $\lambda_{\text{rem}}(t)=0$, so that training first focuses on establishing a stable auxiliary signal. In Stage~1 (alignment), we set $\lambda_{\text{aux}}(t)=0$ while keeping $\lambda_{\text{rem}}(t)=0$, so that training focuses on sufficiency learning before necessity is introduced. In Stage~2 (necessity), once the teacher signal becomes stable, we set $\lambda_{\text{rem}}(t)>0$ and increase $\lambda_{\text{reg}}(t)$ to strengthen necessity learning and regularization.

\section{Experiments}
\subsection{Datasets}
Evaluating explanations in multi-label classification is challenging because real-world graphs rarely provide ground-truth (GT) rationales, and fidelity alone cannot fully capture structural correctness~\cite{christiansen2023faithful,nandan2025graphxai}. We evaluate on five datasets, with standard graph statistics and label-pair statistics summarized in Table~\ref{tab:datasets}. The label-pair statistics characterize the density and direction of label co-occurrence signals available for correlation-aware attribution. The two synthetic datasets are generated by fixed rule-based processes independent of \modelname and provide label-conditioned GT masks for positive $(v,\ell)$ pairs, while the three real-world datasets evaluate prediction and faithfulness without GT explanations. We include the synthetic data generation scripts and configurations in the released code.

\begin{itemize}
\item \textbf{SynAnchor}: a synthetic benchmark for multi-hop reachability evidence. We generate an undirected background graph with random edges. For each label, we sample anchor nodes and add a small number of incident edges to improve positive coverage. A node is positive for label $\ell$ if it reaches any label-$\ell$ anchor within a prescribed hop range. The GT explanation for each positive $(v,\ell)$ is the undirected shortest path from $v$ to its closest label-$\ell$ anchor.

\item \textbf{SynMotif}: a synthetic benchmark for local motif-based evidence. For each label, we allocate a small label-specific support pool and sample positive nodes from the remaining shared nodes. For each positive $(v,\ell)$ pair, we inject a compact label-specific motif by connecting $v$ to support nodes and adding internal edges among them. The GT explanation is exactly the set of injected motif edges for label $\ell$.

\item \textbf{BlogCatalog}: a relatively dense real-world multi-label social network, with labels indicating group or interest memberships.

\item \textbf{YouTube}: a larger and sparser real-world multi-label social network, with labels indicating community or group memberships.

\item \textbf{HumLoc}: a biological multi-label graph for protein subcellular localization, where nodes are proteins, edges are protein-protein interactions, and labels indicate possible subcellular locations.
\end{itemize}

\begin{table}[t]
\centering
\caption{Dataset statistics. $N$, $E$, $d$, and $L$ denote the numbers of nodes, undirected edges, feature dimensions, and labels. Avg. Pair denotes the average number of within-node label pairs per node. Avg. Jac. and Avg. $\rho_Y$ denote the average Jaccard similarity and Spearman correlation between binary label vectors over all label pairs, respectively.}
\scriptsize
\setlength{\tabcolsep}{3.5pt}
\begin{tabular}{l c c c c c c c}
\toprule
Dataset & $N$ & $E$ & $d$ & $L$ & Avg. Pair & Avg. Jac. & Avg. $\rho_Y$ \\
\midrule
SynAnchor   & 3{,}000  & 9{,}476   & 16 & 30 & 2.60 & 0.045 & 0.020 \\
SynMotif    & 3{,}000  & 29{,}849  & 16 & 30 & 3.08 & 0.053 & 0.033 \\
BlogCatalog & 10{,}312 & 667{,}966 & 8  & 39 & 0.62 & 0.010 & -0.009 \\
YouTube     & 22{,}693 & 192{,}722 & 3  & 47 & 1.93 & 0.026 & 0.020 \\
HumLoc      & 3{,}106  & 18{,}496  & 32 & 14 & 0.20 & 0.009 & -0.044 \\
\bottomrule
\end{tabular}
\label{tab:datasets}
\end{table}

\subsection{Baselines}
We compare against three groups of baselines: prediction baselines without explanations, a self-explainable baseline that learns explanations during training, and post-hoc explanation baselines that explain a frozen predictor.

\subsubsection{Prediction baselines}
\begin{itemize}
    \item \textbf{Predict-Only (PO)}: our shared prediction backbone trained without the explanation module.
    \item \textbf{GAT}: a multi-label graph attention network baseline.
    \item \textbf{ML-GCN}~\cite{shi2020multi}: a multi-label GNN baseline that models label correlations through coupled propagation over nodes and labels.
    \item \textbf{LIP}~\cite{sun2025multi}: Predict-Only backbone augmented with label-importance reweighting, where per-label losses are weighted according to label influence.
    \item \textbf{Binary Relevance GNN (BR)}: a binary-relevance baseline that trains one independent classifier per label.
\end{itemize}

\subsubsection{Self-explainable baseline}
\begin{itemize}
\item \textbf{GSAT-style GNN (GSAT)}~\cite{miao2022interpretable}: a stochastic attention GNN adapted to MLNC by replacing the original output layer with sigmoid multi-label outputs and training it with BCE loss. Since GSAT learns a task-level subgraph mask and does not provide a native label-query mechanism, we use the same learned mask for all predicted labels of a node during explanation evaluation.
\end{itemize}

\subsubsection{Explanation baselines}
All post-hoc explainers are built on the same frozen Predict-Only predictor and adapted to MLNC by explaining each predicted positive label independently.
\begin{itemize}
    \item \textbf{Ours-PostHoc (Ours-PH)}: a post-hoc version of our explanation module.
    \item \textbf{GNNExplainer}~\cite{ying2019gnnexplainer}: an optimization-based explainer that learns a soft edge mask for each target label independently.
    \item \textbf{PGExplainer}~\cite{luo2020parameterized}: a parametric explainer that predicts instance-wise edge selection probabilities.
\end{itemize}

\subsection{Experimental Setup}
We use a transductive setting with fixed train/validation/test splits shared by all methods. Models are trained with Adam and early stopping, selected by validation micro- and macro-AUPRC, and evaluated once on the test set. Label imbalance is handled using per-label positive weights computed from the training split. For F1-based reporting, we select a single global threshold on the validation set by grid search over $[0.1,0.9]$ with step size $0.01$ and apply it unchanged to the test set. Baseline hyperparameters follow the original papers or official implementations when available; otherwise, they are selected using the same validation protocol. All final configurations are fixed before test evaluation and included in the released code.

For explanation evaluation, each method explains the predicted positive labels of each test node. Explanations are extracted from the same local $K$-hop computation subgraph with $K=2$. Since message passing uses directed edges, opposite directions of the same connection are merged into one undirected edge group before selection. We use the same adaptive Top-$M$ protocol with bounded minimum and maximum budgets for all methods, ensuring that explanation quality is compared under identical post-processing.

On synthetic datasets, we report F1 and IoU against GT explanation masks. On all datasets, we report Fidelity+ and Fidelity-:
\begin{equation} 
\begin{aligned} \mathrm{Fid}^{+}_{i,c} = 1- \left( p^{\mathrm{keep}}_{i,c} - p^{\mathrm{full}}_{i,c} \right)^2, \qquad \mathrm{Fid}^{-}_{i,c} = p^{\mathrm{full}}_{i,c} - p^{\mathrm{drop}}_{i,c}. 
\end{aligned} 
\end{equation}
Higher $\mathrm{Fid}^{+}$ and $\mathrm{Fid}^{-}$ indicate better sufficiency and necessity, respectively. All prediction and explanation results are averaged over three random seeds. In each run, explanation metrics are computed on 100 test nodes and then averaged.

\begin{table}[t]
\centering
\caption{Node prediction performance. Best results are in bold and second-best results are underlined.}
\label{tab:pred_main_row}
\tiny
\setlength{\tabcolsep}{6pt}
\renewcommand{\arraystretch}{1}
\begin{adjustbox}{max width=\textwidth}
\begin{tabular}{llcccc}
\toprule
Dataset & Method & Micro F1 & Macro F1 & $\mu$AUPRC & MAUPRC \\
\midrule

\multirow{6}{*}{SynAnchor}
& PO   & 0.888$\pm$0.018 & 0.813$\pm$0.042 & 0.935$\pm$0.016 & 0.915$\pm$0.022 \\
& GAT        & \underline{0.893$\pm$0.006} & \underline{0.822$\pm$0.061} & \textbf{0.962$\pm$0.005} & \textbf{0.953$\pm$0.027} \\
& ML-GCN     & 0.302$\pm$0.051 & 0.273$\pm$0.076 & 0.242$\pm$0.070 & 0.322$\pm$0.085 \\
& LIP        & 0.883$\pm$0.033 & 0.773$\pm$0.009 & 0.938$\pm$0.021 & 0.879$\pm$0.030 \\
& BR         & 0.792$\pm$0.026 & 0.661$\pm$0.049 & 0.852$\pm$0.035 & 0.795$\pm$0.049 \\
& GSAT   & 0.524$\pm$0.022 & 0.388$\pm$0.046 & 0.560$\pm$0.013 & 0.481$\pm$0.018 \\
& \textbf{\modelname} & \textbf{0.915$\pm$0.018} & \textbf{0.844$\pm$0.043} & \underline{0.946$\pm$0.014} & \underline{0.930$\pm$0.024} \\
\midrule

\multirow{6}{*}{SynMotif}
& PO   & 0.751$\pm$0.027 & 0.626$\pm$0.048 & 0.714$\pm$0.012 & 0.839$\pm$0.039 \\
& GAT        & 0.528$\pm$0.054 & 0.483$\pm$0.047 & 0.560$\pm$0.056 & \underline{0.933$\pm$0.019} \\
& ML-GCN     & 0.121$\pm$0.011 & 0.109$\pm$0.017 & 0.055$\pm$0.004 & 0.084$\pm$0.010 \\
& LIP        & \underline{0.866$\pm$0.020} & \underline{0.752$\pm$0.043} & \underline{0.871$\pm$0.016} & 0.832$\pm$0.019 \\
& BR         & 0.692$\pm$0.007 & 0.591$\pm$0.034 & 0.670$\pm$0.015 & 0.735$\pm$0.041 \\
& GSAT   &0.736$\pm$0.059 &0.625$\pm$0.031 &0.718$\pm$0.033 & 0.809$\pm$0.005 \\
& \textbf{\modelname} & \textbf{0.903$\pm$0.030} & \textbf{0.779$\pm$0.053} & \textbf{0.900$\pm$0.002} & \textbf{0.942$\pm$0.004} \\

\midrule

\multirow{6}{*}{BlogCatalog}
& PO  & 0.204$\pm$0.033 & 0.076$\pm$0.003 & 0.159$\pm$0.035 & 0.140$\pm$0.013 \\
& GAT        & 0.254$\pm$0.067 & 0.081$\pm$0.021 & 0.217$\pm$0.038 & \underline{0.163$\pm$0.016}\\
& ML-GCN     & 0.253$\pm$0.026 & \textbf{0.101$\pm$0.015} & 0.194$\pm$0.019 & \textbf{0.198$\pm$0.022} \\
& LIP        & \underline{0.290$\pm$0.025} & 0.079$\pm$0.011 & 0.227$\pm$0.039 & 0.126$\pm$0.018 \\
& BR         & 0.243$\pm$0.028 & 0.088$\pm$0.021 & \underline{0.274$\pm$0.052} & 0.121$\pm$0.009 \\
& GSAT   &0.270$\pm$0.006 & 0.077$\pm$0.016 & 0.254$\pm$0.017 & 0.141$\pm$0.020 \\
& \textbf{\modelname} & \textbf{0.404$\pm$0.033} & \underline{0.092$\pm$0.010} & \textbf{0.367$\pm$0.051} & 0.143$\pm$0.025 \\
\midrule

\multirow{6}{*}{YouTube}
& PO   & 0.249$\pm$0.009 & 0.214$\pm$0.019 & 0.242$\pm$0.017 & 0.252$\pm$0.028 \\
& GAT        & 0.224$\pm$0.008 & 0.199$\pm$0.020 & 0.187$\pm$0.009 & 0.212$\pm$0.009 \\
& ML-GCN     & \textbf{0.338$\pm$0.030} & 0.178$\pm$0.010 & 0.219$\pm$0.010 & 0.196$\pm$0.061 \\
& LIP        & 0.253$\pm$0.023 & 0.226$\pm$0.015 & 0.243$\pm$0.019 & \textbf{0.267$\pm$0.029} \\
& BR         & 0.283$\pm$0.016 & 0.171$\pm$0.009 & 0.231$\pm$0.029 & 0.204$\pm$0.011 \\
& GSAT   &0.275$\pm$0.032 & \underline{0.231$\pm$0.019} & \underline{0.246$\pm$0.015} & 0.253$\pm$0.026 \\
& \textbf{\modelname} & \underline{0.287$\pm$0.023} & \textbf{0.232$\pm$0.022} & \textbf{0.252$\pm$0.023} & \underline{0.255$\pm$0.031} \\
\midrule

\multirow{6}{*}{HumLoc}
& PO   &0.537$\pm$0.043 & 0.159$\pm$0.024 & 0.541$\pm$0.053 & \underline{0.299$\pm$0.098} \\
& GAT        &0.579$\pm$0.014 & 0.154$\pm$0.012 & 0.602$\pm$0.035 & 0.296$\pm$0.097 \\
& ML-GCN     &0.494$\pm$0.041 & 0.143$\pm$0.013 & 0.407$\pm$0.047 & 0.272$\pm$0.083 \\
& LIP       &0.596$\pm$0.022 & 0.156$\pm$0.002 & 0.623$\pm$0.035 & 0.212$\pm$0.037 \\
& BR         &0.512$\pm$0.023 & \underline{0.160$\pm$0.011} & 0.428$\pm$0.100 & 0.194$\pm$0.027 \\
& GSAT   &\underline{0.602$\pm$0.008} & 0.146$\pm$0.009 & \textbf{0.639$\pm$0.017} & 0.297$\pm$0.089 \\
& \textbf{\modelname} &\textbf{0.615$\pm$0.041} & \textbf{0.177$\pm$0.028} & \underline{0.625$\pm$0.043} & \textbf{0.318$\pm$0.087} \\

\bottomrule
\end{tabular}
\end{adjustbox}
\end{table}

\subsection{Experimental Results}
Our experiments are designed to evaluate three central questions: whether \modelname maintains strong multi-label prediction performance, whether it produces faithful and compact label-conditioned explanations, and whether label-correlation modeling makes explanation sharing more consistent with label-relatedness.

\subsubsection{Prediction under Correlated Label Evidence}

Table~\ref{tab:pred_main_row} reports the predictive performance on synthetic and real-world multi-label graphs. Overall, \modelname achieves competitive or superior prediction performance, with the clearest gains on the two synthetic datasets. On SynAnchor and SynMotif, where label-relevant structural evidence is explicitly controlled, \modelname obtains strong F1 and AUPRC results. This suggests that jointly learning prediction and explanation does not sacrifice classification accuracy, and can benefit settings where different labels are supported by identifiable graph structures.

On real-world datasets, the gains are more selective. \modelname performs strongly on BlogCatalog at the micro level, while on YouTube it shows stronger macro-level performance and remains competitive in micro-level evaluation. On HumLoc, \modelname improves F1 while maintaining competitive AUPRC. These results indicate that the benefit of \modelname's joint label-aware design depends on the dataset-specific graph and label structure. Rather than claiming uniform dominance across all metrics, the prediction results show that \modelname provides a reliable predictive basis for correlated evidence attribution.

\begin{table*}[t]
\centering
\caption{Explanation performance. SP denotes sparsity. Best results are in bold, and second-best results are underlined.}
\label{tab:exp_main_row}
\scriptsize
\setlength{\tabcolsep}{6pt}
\renewcommand{\arraystretch}{1}
\begin{adjustbox}{max width=\textwidth}
\begin{tabular}{llcccc}
\toprule
Dataset & Method & \(\mathrm{Fid}^{+}\) & \(\mathrm{Fid}^{-}\)  & GT F1 & GT IoU\\
\midrule

\multirow{4}{*}{\shortstack{SynAnchor\\(SP=0.834$\pm$0.009)}}
& Ours-PH  & 0.736$\pm$0.046 & 0.176$\pm$0.026  & 0.095$\pm$0.066 & 0.056$\pm$0.045 \\
& GNNExplainer   & 0.830$\pm$0.013 & \underline{0.269$\pm$0.019}  & 0.163$\pm$0.013 & 0.102$\pm$0.006\\
& PGExplainer    & 0.625$\pm$0.014 & 0.059$\pm$0.018  & 0.211$\pm$0.030 & 0.152$\pm$0.024\\
& GSAT   & \underline{0.836$\pm$0.017} & 0.065$\pm$0.006 & \underline{0.311$\pm$0.016} & \underline{0.204$\pm$0.016} \\
& \textbf{\modelname} & \textbf{0.860$\pm$0.023} & \textbf{0.383$\pm$0.017}  & \textbf{0.463$\pm$0.007} & \textbf{0.337$\pm$0.006}\\
\midrule

\multirow{4}{*}{\shortstack{SynMotif\\(SP=0.897$\pm$0.001)}}
& Ours-PH  & 0.825$\pm$0.069 & -0.151$\pm$0.022 & \underline{0.210$\pm$0.022} & \underline{0.118$\pm$0.013}\\
& GNNExplainer   & 0.389$\pm$0.024 & \underline{0.116$\pm$0.011}  & 0.191$\pm$0.003 & 0.112$\pm$0.002\\
& PGExplainer   & 0.777$\pm$0.014 & -0.046$\pm$0.013  & 0.122$\pm$0.017 & 0.073$\pm$0.011\\
& GSAT   &\textbf{0.926$\pm$0.026} & 0.001$\pm$0.023 & 0.152$\pm$0.008 & 0.084$\pm$0.005 \\
& \textbf{\modelname}  & \underline{0.878$\pm$0.071} & \textbf{0.157$\pm$0.016}  & \textbf{0.320$\pm$0.009} & \textbf{0.193$\pm$0.006}\\
\midrule

\multirow{4}{*}{\shortstack{BlogCatalog\\(SP=0.977$\pm$0.007)}}
& Ours-PH  & 0.742$\pm$0.003 & 0.131$\pm$0.008  & -- & --\\
& GNNExplainer   & 0.782$\pm$0.000 & \textbf{0.255$\pm$0.000}  & -- & --\\
& PGExplainer    & 0.827$\pm$0.003 & 0.037$\pm$0.010  & -- & --\\
& GSAT   &\textbf{0.921$\pm$0.007} & 0.060$\pm$0.009 & -- & -- \\
& \textbf{\modelname}  & \underline{0.853$\pm$0.013} & \underline{0.144$\pm$0.006}  & -- & --\\
\midrule

\multirow{4}{*}{\shortstack{YouTube\\(SP=0.787$\pm$0.031)}}
& Ours-PH  & 0.880$\pm$0.005 & 0.065$\pm$0.002  & -- & --\\
& GNNExplainer   & 0.879$\pm$0.012 & \underline{0.163$\pm$0.004}  & -- & --\\
& PGExplainer    & 0.879$\pm$0.001 & 0.051$\pm$0.028  & -- & --\\
& GSAT   & \underline{0.901$\pm$0.005} & 0.093$\pm$0.017 & -- & -- \\
& \textbf{\modelname}  & \textbf{0.924$\pm$0.005} & \textbf{0.202$\pm$0.005}  & -- & --\\
\midrule

\multirow{4}{*}{\shortstack{HumLoc\\(SP=0.857$\pm$0.010)}}
& Ours-PH  & \underline{0.953$\pm$0.013} & \textbf{0.315$\pm$0.060}  & -- & --\\
& GNNExplainer   & 0.863$\pm$0.019 & 0.228$\pm$0.024  & -- & --\\
& PGExplainer    & 0.733$\pm$0.021 & 0.055$\pm$0.022  & -- & --\\
& GSAT   &0.908$\pm$0.009 & 0.042$\pm$0.002 & -- & -- \\
& \textbf{\modelname}  & \textbf{0.962$\pm$0.017} & \underline{0.264$\pm$0.055}  & -- & --\\

\bottomrule
\end{tabular}
\end{adjustbox}
\end{table*}

\subsubsection{Faithfulness of Label-Conditioned Evidence Attribution}

Table~\ref{tab:exp_main_row} evaluates whether the selected edges provide faithful evidence for each predicted label. On the synthetic datasets, where GT rationales are available, \modelname achieves the best GT F1 and GT IoU on both SynAnchor and SynMotif. This is important for correlated evidence attribution because the explanations are evaluated not only by their effect on the predictor, but also by whether they recover the label-conditioned structural rationales used to generate the labels.

In terms of fidelity, \modelname also shows a strong overall balance between sufficiency and necessity. On SynAnchor, it achieves the best $\mathrm{Fid}^{+}$ and $\mathrm{Fid}^{-}$, indicating that the selected evidence both preserves the prediction and becomes decision-critical when removed. On SynMotif, GSAT obtains the highest $\mathrm{Fid}^{+}$, but \modelname achieves the highest $\mathrm{Fid}^{-}$ and substantially better GT alignment, suggesting that high sufficiency alone does not necessarily imply structurally correct label-specific attribution. On real-world datasets, where GT rationales are unavailable, \modelname remains competitive or strong in fidelity: it achieves the best $\mathrm{Fid}^{+}$ and $\mathrm{Fid}^{-}$ on YouTube, the best $\mathrm{Fid}^{+}$ and second-best $\mathrm{Fid}^{-}$ on HumLoc, and a balanced second-best performance on BlogCatalog. These results suggest that \modelname does not merely select compact subgraphs, but identifies edges that are useful for preserving and perturbing label-specific predictions.

The comparison with baselines further clarifies the role of end-to-end correlation-aware attribution. Post-hoc methods, such as GNNExplainer and PGExplainer, explain a frozen predictor and can be adapted to MLNC by explaining each positive label independently. However, they do not explicitly model whether evidence should be shared or separated according to label relationships. GSAT jointly learns prediction and explanation, but its learned mask is task-level rather than label-conditioned, making it difficult to distinguish evidence for positively related labels from evidence for weakly or negatively associated labels. In contrast, \modelname learns edge attribution during training and conditions explanations on correlation-aware label representations, making it better suited to MLNC where labels may require distinct but partially shared rationales.

\subsubsection{Qualitative Analysis of Label-Specific and Shared Evidence}

\begin{figure*}[t]
  \centering
  \includegraphics[width=0.9\textwidth]{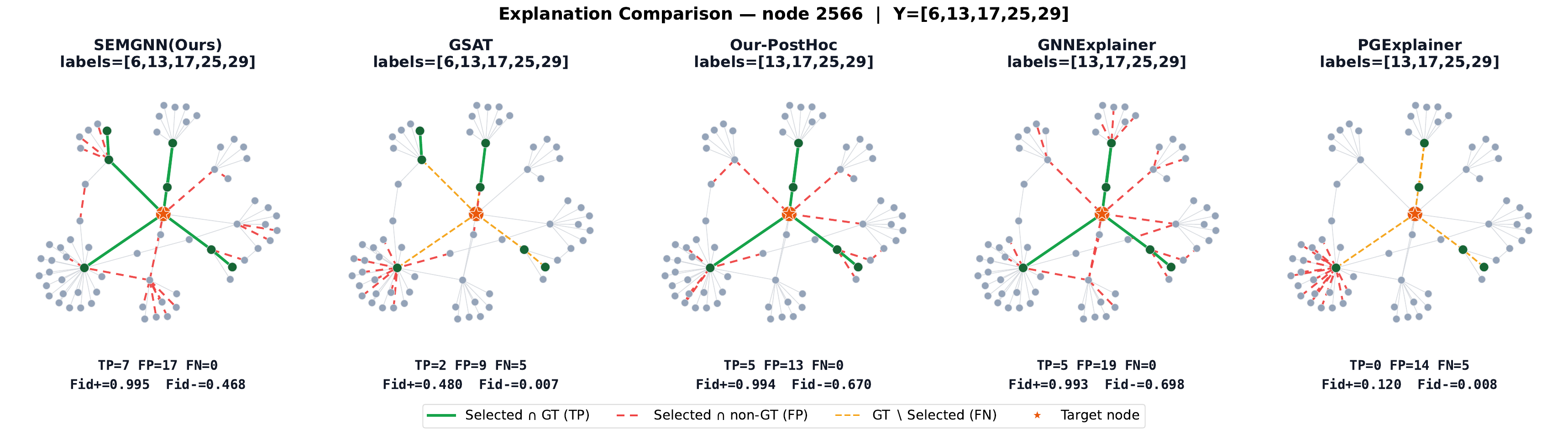}
  \caption{Visualization of multi-label explanations for a single node on SynAnchor. $Y$ denotes the ground-truth label set of the target node, while \texttt{labels=[...]} denotes the predicted positive labels explained by each method. Green solid edges denote selected GT edges, i.e., true positives (TP); red dashed edges denote selected non-GT edges, i.e., false positives (FP); orange dashed edges denote GT edges missed by the explanation, i.e., false negatives (FN); and the orange star marks the target node. Each panel reports TP/FP/FN counts and Fid+/Fid- scores.}
  \label{fig:v5visual}
\end{figure*}

Fig.~\ref{fig:v5visual} visualizes a multi-label explanation on SynAnchor. In this example, \modelname recovers the complete label set of the target node and identifies all visible GT explanatory edges without missing true rationale edges. Although GSAT also predicts the same label set, its shared task-level mask misses several GT rationale edges, suggesting that self-explainable training alone may be insufficient for label-conditioned evidence attribution. The post-hoc baselines miss one positive label in this case and only partially recover the corresponding structural rationale. Although a single qualitative example is not conclusive by itself, it is consistent with the quantitative GT F1/IoU results in Table~\ref{tab:exp_main_row} and illustrates the intended behavior of \modelname: multi-label explanations should attribute each predicted label to its supporting structural evidence, rather than producing a generic subgraph explanation.

\begin{figure*}[t]
  \centering
  \includegraphics[width=0.9\textwidth]{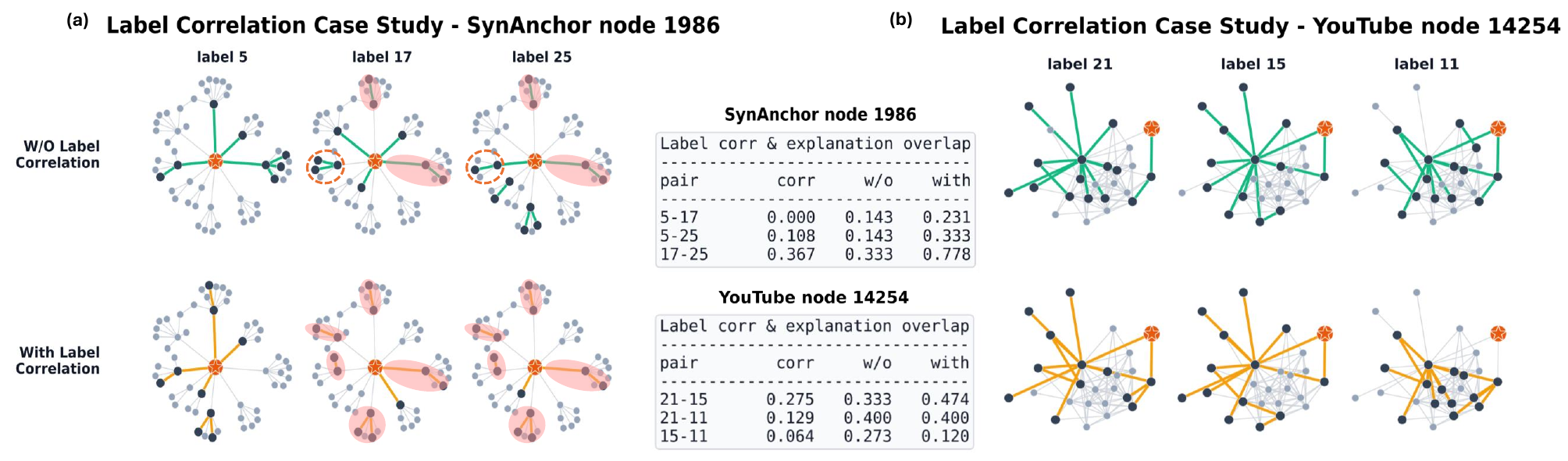}
    \caption{Case studies of label-correlation-guided explanation sharing on SynAnchor (a) and YouTube (b). For each case, the top row shows label-wise explanations without label correlation (green edges), and the bottom row shows explanations with label correlation (orange edges). The center tables report pairwise label correlation and explanation overlap before and after enabling label correlation. In the SynAnchor case, structures shared by labels 17 and 25 are highlighted in red, while partially overlapping structures are marked with dashed circles. For the denser YouTube case, overlap changes are summarized in the center table for readability.}
    \label{fig:label}
\end{figure*}

Fig.~\ref{fig:label} further examines how label correlation affects evidence sharing through two case studies on SynAnchor and YouTube. In the SynAnchor example, labels 17 and 25 form the most strongly correlated pair among the queried labels; after enabling label correlation, their shared edge groups increase from 2 to 5, corresponding to an explanation-overlap increase from 0.333 to 0.778. In the YouTube example, the strongly correlated pair also shows a clear increase in explanation overlap after enabling label correlation, as summarized in the center table. Importantly, the effect is not a uniform increase across all label pairs: pairs with weaker correlations show smaller or even reduced overlap. This suggests that label correlation provides a structured prior for organizing shared evidence, rather than collapsing all label-wise explanations into one common subgraph. In other words, \modelname uses label relationships to modulate evidence sharing: related labels may share more support, while less related labels can retain distinct explanatory structures.

\begin{table*}[t]
\centering
\caption{Ablation study on node prediction and explanation performance. Best results are in bold, and second-best results are underlined.}
\label{tab:ablation}
\scriptsize
\setlength{\tabcolsep}{2.5pt}
\renewcommand{\arraystretch}{1.05}
\begin{tabular*}{\textwidth}{@{\extracolsep{\fill}}llcccccccc}
\toprule
\multirow{2}{*}{Dataset} & \multirow{2}{*}{Variant}
& \multicolumn{4}{c}{Prediction}
& \multicolumn{4}{c}{Explanation} \\
\cmidrule(lr){3-6} \cmidrule(lr){7-10}
& & Micro F1 & Macro F1 & $\mu$AUPRC & MAUPRC
& $\mathrm{Fid}^{+}$ & $\mathrm{Fid}^{-}$ & GT F1 & GT IoU \\
\midrule

\multirow{4}{*}{\shortstack{SynAnchor}}
& \textbf{\modelname}
& \textbf{0.915$\pm$0.018} & \textbf{0.844$\pm$0.043} & \textbf{0.946$\pm$0.014} & \underline{0.930$\pm$0.024}
& 0.860$\pm$0.023 & \textbf{0.383$\pm$0.017} & \underline{0.463$\pm$0.007} & \underline{0.337$\pm$0.006} \\
& w/o Pred-L
& 0.905$\pm$0.032 & \underline{0.823$\pm$0.005} & 0.945$\pm$0.022 & 0.916$\pm$0.040
& \underline{0.866$\pm$0.006} & 0.359$\pm$0.028 & \textbf{0.466$\pm$0.023} & \textbf{0.338$\pm$0.018} \\
& w/o Expl-L
& \underline{0.910$\pm$0.023} & 0.821$\pm$0.021 & \underline{0.945$\pm$0.023} & \textbf{0.935$\pm$0.028}
& \textbf{0.879$\pm$0.028} & \underline{0.371$\pm$0.008} & 0.456$\pm$0.006 & 0.332$\pm$0.005 \\
& Base
& 0.899$\pm$0.023 & 0.814$\pm$0.023 & 0.939$\pm$0.019 & 0.919$\pm$0.035
& 0.861$\pm$0.012 & 0.354$\pm$0.011 & 0.448$\pm$0.026 & 0.325$\pm$0.019 \\
\midrule

\multirow{4}{*}{\shortstack{SynMotif}}
& \textbf{\modelname}
& \textbf{0.903$\pm$0.030} & \textbf{0.779$\pm$0.053} & \textbf{0.900$\pm$0.002} & \textbf{0.942$\pm$0.004}
& \textbf{0.878$\pm$0.071} & \textbf{0.157$\pm$0.016} & \underline{0.320$\pm$0.009} & \underline{0.193$\pm$0.006} \\
& w/o Pred-L
& 0.881$\pm$0.033 & 0.744$\pm$0.065 & 0.896$\pm$0.012 & 0.921$\pm$0.014
& \underline{0.851$\pm$0.084} & \underline{0.130$\pm$0.013} & 0.297$\pm$0.016 & 0.181$\pm$0.009 \\
& w/o Expl-L
& \underline{0.901$\pm$0.011} & \underline{0.773$\pm$0.024} & \underline{0.897$\pm$0.009} & \underline{0.941$\pm$0.011}
& 0.749$\pm$0.095 & 0.108$\pm$0.009 & 0.291$\pm$0.016 & 0.176$\pm$0.009 \\
& Base
& 0.875$\pm$0.029 & 0.741$\pm$0.032 & 0.890$\pm$0.021 & 0.933$\pm$0.020
& 0.673$\pm$0.115 & 0.096$\pm$0.016 & \textbf{0.331$\pm$0.013} & \textbf{0.201$\pm$0.009} \\
\midrule

\multirow{4}{*}{\shortstack{BlogCatalog}}
& \textbf{\modelname}
& \textbf{0.404$\pm$0.033} & \textbf{0.092$\pm$0.010} & \textbf{0.367$\pm$0.051} & \textbf{0.143$\pm$0.025}
& 0.853$\pm$0.013 & \textbf{0.144$\pm$0.006} & -- & -- \\
& w/o Pred-L
& 0.354$\pm$0.040 & 0.077$\pm$0.012 & 0.324$\pm$0.060 & 0.135$\pm$0.025
& \underline{0.855$\pm$0.013} & \underline{0.123$\pm$0.007} & -- & -- \\
& w/o Expl-L
& 0.370$\pm$0.031 & \underline{0.091$\pm$0.023} & \underline{0.348$\pm$0.039} & \underline{0.142$\pm$0.024}
& \textbf{0.860$\pm$0.003} & 0.122$\pm$0.009 & -- & -- \\
& Base
& \underline{0.370$\pm$0.046} & 0.081$\pm$0.009 & 0.342$\pm$0.063 & 0.128$\pm$0.014
& 0.841$\pm$0.008 & 0.119$\pm$0.004 & -- & -- \\
\midrule

\multirow{4}{*}{\shortstack{YouTube}}
& \textbf{\modelname}
& \underline{0.287$\pm$0.023} & \underline{0.232$\pm$0.022} & \underline{0.252$\pm$0.023} & \textbf{0.255$\pm$0.031}
& \textbf{0.924$\pm$0.005} & \underline{0.202$\pm$0.005} & -- & -- \\
& w/o Pred-L
& 0.284$\pm$0.018 & 0.227$\pm$0.014 & 0.251$\pm$0.022 & 0.240$\pm$0.032
& \underline{0.887$\pm$0.019} & \textbf{0.203$\pm$0.006} & -- & -- \\
& w/o Expl-L
& \textbf{0.288$\pm$0.016} & 0.222$\pm$0.017 & \textbf{0.257$\pm$0.024} & 0.243$\pm$0.027
& 0.874$\pm$0.023 & 0.193$\pm$0.012 & -- & -- \\
& Base
& 0.276$\pm$0.011 & \textbf{0.233$\pm$0.023} & 0.250$\pm$0.018 & \underline{0.245$\pm$0.023}
& 0.881$\pm$0.023 & 0.197$\pm$0.003 & -- & -- \\
\midrule

\multirow{4}{*}{\shortstack{HumLoc}}
& \textbf{\modelname} &0.615$\pm$0.041 & \textbf{0.177$\pm$0.028} & 0.625$\pm$0.043 & \underline{0.318$\pm$0.087}
& \textbf{0.962$\pm$0.017} & \textbf{0.264$\pm$0.055} & -- & -- \\
& w/o Pred-L
& \underline{0.617$\pm$0.032} & \underline{0.163$\pm$0.010} & \textbf{0.663$\pm$0.016} & \textbf{0.321$\pm$0.079}
& \underline{0.956$\pm$0.009} & \underline{0.262$\pm$0.044} & -- & -- \\
& w/o Expl-L
& \textbf{0.626$\pm$0.018} & 0.163$\pm$0.002 & \underline{0.638$\pm$0.027} & 0.311$\pm$0.092
& 0.956$\pm$0.003 & 0.254$\pm$0.016 & -- & -- \\
& Base
& 0.596$\pm$0.032 & 0.159$\pm$0.011 & 0.633$\pm$0.036 & 0.316$\pm$0.097
& 0.951$\pm$0.009 & 0.242$\pm$0.015 & -- & -- \\
\bottomrule
\end{tabular*}
\end{table*}

\subsubsection{Ablation Study}

Table~\ref{tab:ablation} studies how prediction-side and explanation-side label modeling contribute to \modelname. We compare the full model with three variants: removing the prediction-side label-correlation residual (w/o Pred-L), removing the explanation-side label-aware scorer (w/o Expl-L), and removing both label-aware components (Base).

For prediction, removing the prediction-side label-correlation residual generally leads to lower performance, especially on SynAnchor, SynMotif, and BlogCatalog. This indicates that label dependency is useful for MLNC, where correlated labels can provide complementary decision signals. The effect is less uniform on YouTube and HumLoc, where some AUPRC scores are comparable or slightly higher for ablated variants. This suggests that the predictive contribution of label correlation is dataset-dependent, and avoids over-interpreting label modeling as universally beneficial across all metrics.

For explanation, the full model often provides the strongest sufficiency--necessity balance, particularly in $\mathrm{Fid}^{-}$ on SynAnchor, SynMotif, BlogCatalog, and HumLoc. This indicates that the attributed edges are more likely to be decision-critical when removed. On the synthetic datasets, the full model also achieves strong GT alignment, although some ablated variants can obtain slightly higher GT F1/IoU in isolated cases. This is expected because GT recovery measures agreement with externally defined rationales, while fidelity measures faithfulness to the learned predictor. The ablation results, therefore, suggest that prediction-side and explanation-side label modeling play complementary roles: prediction-side correlation helps learn label-dependent decision signals, while explanation-side label-aware scoring helps convert these signals into label-conditioned edge attribution.



\begin{figure}[t]
    \centering
    \includegraphics[width=0.99\linewidth]{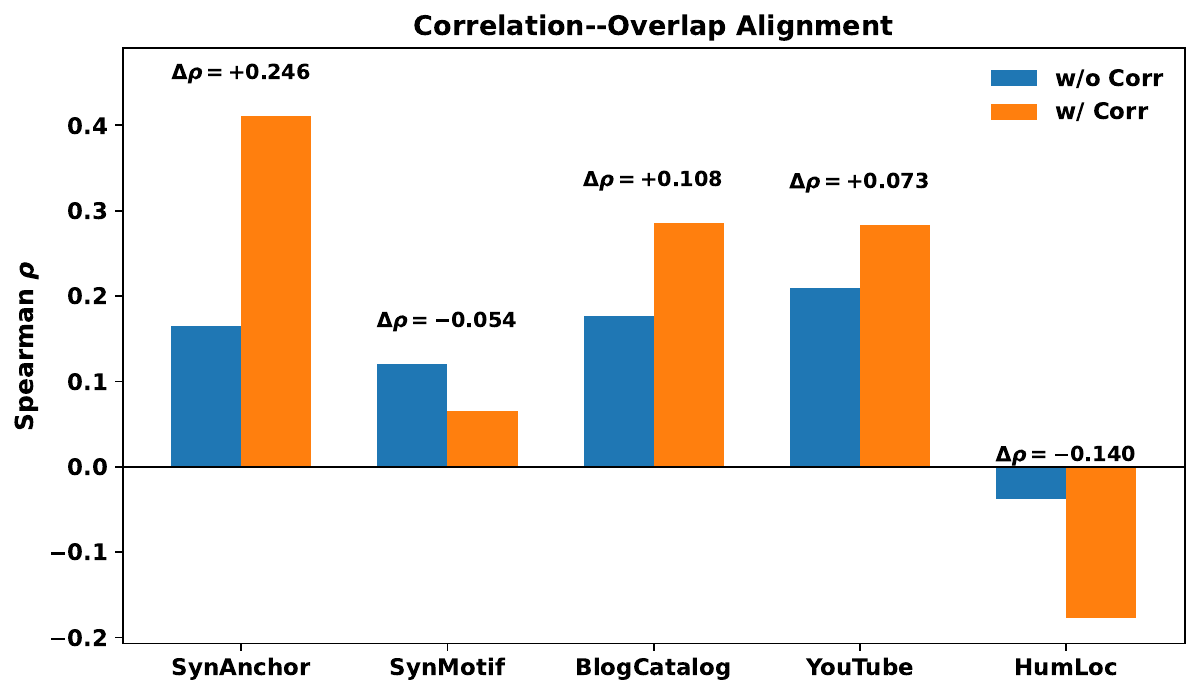}
    \caption{Correlation--overlap alignment before and after enabling label-correlation modeling. The $y$-axis reports Spearman $\rho$ between training-label correlation and label-wise explanation overlap. A larger $\rho$ indicates that more correlated labels tend to share more explanatory evidence. $\Delta\rho$ denotes the change from w/o Corr to w/ Corr. The HumLoc dataset has a negative $\Delta\rho$ because its labels are negatively correlated on average, meaning \modelname~ can adapt to label correlation (positive or negative) to find correlated evidence.}
    \label{fig:label_overlap}
\end{figure}

\subsubsection{Label Correlation and Evidence Sharing}
\label{sec:label_correlation_analysis}
In correlated evidence attribution, label correlation should not force all labels to share the same explanation. Instead, it should encourage evidence sharing when labels are statistically related, while allowing weakly related labels to retain distinct rationales. We construct the label-correlation graph only from the training split to avoid validation or test label leakage. The graph is based on co-occurrence statistics from the multi-hot training label matrix, and is sparsified, augmented with self-loops, and normalized before label-aware propagation.


Fig.~\ref{fig:label_overlap} further examines whether label-correlation modeling makes explanation sharing more consistent with label-relatedness. For each comparable label pair, we compute the Spearman correlation $\rho$ between its training-label correlation and its label-wise explanation overlap. A larger $\rho$ indicates that more correlated labels tend to share more explanatory evidence. We compare \modelname without label-correlation modeling and the full \modelname with label-correlation modeling. The results show that enabling label correlation increases correlation--overlap alignment on SynAnchor, BlogCatalog, and YouTube. The improvement is most pronounced on SynAnchor, where $\rho$ increases from 0.165 to 0.410, indicating that shared explanations become substantially more aligned with label relatedness. BlogCatalog and YouTube also show positive alignment gains, suggesting that the label-correlation prior helps organize shared evidence in real-world social graphs. On SynMotif, $\rho$ decreases slightly, which is consistent with its motif-specific construction, where different labels are expected to rely on distinct local rationales rather than shared structural evidence. On HumLoc, $\rho$ decreases and becomes more negative after label-correlation modeling. This result should be interpreted cautiously but is also informative: as shown in Table~\ref{tab:datasets}, HumLoc has the smallest label space, the lowest average number of within-node label pairs, the lowest average label-pair Jaccard similarity, and the most negative average label-pair Spearman correlation. These statistics indicate sparse and weak positive label co-occurrence signals. In such cases, desirable behavior is not to force more overlap, but to preserve distinct label-conditioned evidence when labels are weakly or negatively associated. This highlights an advantage of \modelname over post-hoc or task-level explainers: label correlations provide a structured prior for modulating evidence sharing, rather than uniformly increasing explanation overlap.

Taken together, the ablation and correlation--overlap alignment analyses show that label information contributes to \modelname in two complementary ways. First, prediction-side label modeling improves or maintains multi-label prediction by incorporating label dependency into the decision function. Second, explanation-side label modeling organizes label-conditioned edge attribution by modulating how evidence is shared or separated across related labels. These findings support the main claim of correlated evidence attribution: \modelname attributes predictions to label-specific evidence while using label correlations as a structured prior for shared or distinct explanatory support.

\section{Conclusion}
This paper studies correlated evidence attribution for multi-label graph learning, where explanations should identify label-specific structural evidence while adapting evidence sharing to label relationships. We propose \modelname, an end-to-end self-explainable multi-label GNN that jointly learns node prediction and edge-level attribution under a unified objective. By incorporating label correlations into both prediction and explanation, \modelname produces label-conditioned explanations that can share support for positively related labels while preserving distinct rationales when labels are weakly or even negatively associated. Experiments on synthetic and real-world graphs show that \modelname achieves competitive predictive performance and provides faithful explanations with a strong sufficiency--necessity balance. Further qualitative and correlation analyses show that label correlations help underpin structures with much more insightful explanations than collapsing all labels into a single explanation.



\bibliographystyle{IEEEtran}
\bibliography{mybibliography}

\end{document}